\documentclass[12pt]{article}

\usepackage[utf8]{inputenc}
\usepackage[margin=1in]{geometry}
\usepackage{amsmath,amssymb}
\usepackage{graphicx}
\usepackage{booktabs}
\usepackage{array}
\usepackage{tabularx}
\usepackage{enumitem}
\usepackage{float}
\usepackage[section]{placeins}
\usepackage{titling}
\usepackage[numbers]{natbib}
\usepackage{xcolor}
\usepackage{url}

\usepackage[
    colorlinks=true,
    linkcolor=blue,
    citecolor=blue,
    urlcolor=blue
]{hyperref}

\newcolumntype{Y}{>{\raggedright\arraybackslash}X}

\title{\large\textbf{AI Agents Do Not Fail Alone: The Context Fails First}}

\author{
\begin{tabular}{c}
\normalsize Fouad Bousetouane\textsuperscript{1,2} \tabularnewline
\small \textsuperscript{1}\href{https://www.proofagent.ai}{ProofAgent.ai} \tabularnewline
\small \textsuperscript{2}The University of Chicago, USA \tabularnewline
\footnotesize \href{mailto:bousetouane@uchicago.edu}{\nolinkurl{bousetouane@uchicago.edu}}
\end{tabular}
}

\date{}

\begin{document}

\maketitle

\begin{abstract}
Context engineering has become central to building reliable AI agents, yet it remains largely unmeasured. Agents do not fail in isolation: their behavior is shaped by the instructions, tools, memory, retrieved knowledge, guardrails, and untrusted inputs accumulated in their context. When this context is weak, agents drift, hallucinate, misuse tools, ignore constraints, become vulnerable to injection, and waste tokens.

This paper validates context-engineering quality as an independent leading indicator of agent reliability. We implement the measurement in ProofAgent-Harness\footnote{\url{https://github.com/ProofAgent-ai/proofagent-harness}}, an open-source infrastructure for AI agent evaluation that uses multi-juror, consensus-based scoring. The harness assesses context across seven criteria: role clarity, guardrail coverage, instruction consistency, tool schema quality, grounding sufficiency, injection hardening, and token efficiency. Crucially, the context score is isolated from behavioral metrics and release decisions, enabling a non-circular validation.

Through a controlled context-quality study across regulated agent domains, holding the model fixed and varying only the context, we show that context-quality criteria consistently predict their corresponding behavioral outcomes. Grounding predicts hallucination resistance, guardrails predict manipulation resistance, instruction consistency predicts instruction following, and tool-schema quality predicts tool use. These findings establish context measurement as a validated preflight signal for agent reliability and position context engineering as an auditable layer of agent evaluation and governance.
\end{abstract}

\newpage
\tableofcontents
\newpage

% ============================================================
\section{Introduction}
\label{sec:introduction}

AI agents do not fail alone. Their behavior is shaped by the context in which they operate: system instructions, tool schemas, retrieved knowledge, memory, prior turns, guardrails, and untrusted external inputs. As agents move from single-turn assistants to multi-step systems that call tools, write artifacts, retain state, and act across workflows, context becomes a hidden reliability layer. When this layer is poorly engineered, agents can drift from their role, hallucinate unsupported facts, misuse tools, follow conflicting instructions, become vulnerable to prompt injection, or waste tokens on irrelevant information.

This shift has led practitioners to distinguish \emph{context engineering} from prompt engineering. Prompt engineering focuses on crafting or refining a single instruction. Context engineering concerns the full information environment supplied to the model: which instructions are present, how tools are described, what knowledge is grounded, how memory is represented, how trusted and untrusted content are separated, and how the working context evolves across turns. Prior work on long-context behavior shows that adding more context does not guarantee reliable use of information \cite{liu2024lost}. Retrieval-augmented generation demonstrates the value of grounding model outputs in external knowledge \cite{lewis2020rag}. Recent agent-building guidance similarly emphasizes that reliable agents require careful management of context, tools, and external information rather than prompt wording alone \cite{anthropic2025context}.

Despite its importance, context engineering remains largely unmeasured. Teams often inspect prompts manually, debug failures after deployment, or rely on behavioral evaluations that reveal whether an agent failed without isolating whether the failure originated in the context itself. This creates an operational blind spot. A full adversarial evaluation may show that an agent hallucinated, ignored a policy, or misused a tool, but it may not explain whether the upstream cause was weak grounding, unclear role definition, conflicting instructions, underspecified tool schemas, missing guardrails, or poor separation between trusted instructions and untrusted inputs.

This paper validates context-engineering quality as an independent leading indicator of AI agent reliability. We define context quality as a seven-criterion construct covering role clarity, guardrail coverage, instruction consistency, tool schema quality, grounding sufficiency, injection hardening, and token efficiency. The construct is implemented in ProofAgent-Harness, an open-source infrastructure for adversarial AI agent evaluation \cite{bousetouane2026proofagent}. ProofAgent-Harness operationalizes Human-on-the-Bridge (HOB) evaluation through multi-juror, consensus-based scoring \cite{bousetouane2026hob}, producing both behavioral scores and an isolated context-quality assessment.

The key design requirement is isolation. The context score does not enter the behavioral metrics, final grade, or release decision. It grades the setup in which the agent reasons, not the behavior being scored. This separation enables a non-circular validation: if context-quality criteria predict downstream behavioral outcomes while remaining independent of the behavioral score, then context quality functions as a genuine reliability signal rather than a restatement of the final evaluation.

We test this hypothesis through a controlled context-quality study across regulated agent domains. Holding the model fixed, we vary only the context across three levels: poor, structured, and hardened. The poor context contains vague role definition, weak or missing tool guidance, limited grounding, and little protection against untrusted inputs. The structured context adds clearer role boundaries, typed tools, grounding, and more efficient organization. The hardened context adds stronger guardrails, escalation behavior, injection separation, and confirm-before-mutate patterns. This controlled context ladder allows us to test whether improvements in measured context quality correspond to improvements in independent behavioral outcomes.

The results support the validation claim. Context-quality criteria consistently predict their corresponding behavioral signals: grounding sufficiency predicts hallucination resistance, guardrail coverage predicts manipulation resistance, instruction consistency predicts instruction following, and tool schema quality predicts tool use. The artifact study further shows that grounded context improves deliverable reliability, extending the signal beyond interactive dialogue. The results also show that the cost of weak context is not primarily token cost: the weakest context can be the cheapest per call while producing the most dangerous behavior.

This paper makes three contributions:

\begin{enumerate}
    \item We define context-engineering quality as a measurable seven-criterion construct for AI agents.
    \item We implement this construct in ProofAgent-Harness using multi-juror, consensus-based scoring while keeping the context score isolated from behavioral metrics and release decisions.
    \item We validate that context-quality criteria predict independent downstream behavioral outcomes across controlled context variations in regulated agent domains.
\end{enumerate}

The remainder of the paper is organized as follows. Section~2 summarizes related work on context engineering, long-context degradation, grounding, prompt injection, and agent evaluation. Section~3 defines the seven-criterion context-quality measurement implemented in ProofAgent-Harness. Section~4 presents the experimental validation, including the study design, results, artifact study, token-cost analysis, and interpretation. Section~5 concludes with the implications of treating context engineering as a measurable layer of agent evaluation and governance.

% ============================================================
\section{Background: From Prompt Engineering to Context Engineering}
\label{sec:background}

The introduction framed agent reliability as a property of both the model and the information environment in which the model reasons. This section develops that claim by situating context engineering within prior work on prompting, retrieval, tool use, memory, compression, injection defense, and multi-agent systems. The goal is not to provide an exhaustive survey, but to identify the main techniques that shape agent context and motivate why their quality should be measured.

We use \emph{context engineering} to mean the design and runtime management of the information supplied to a model during inference. This includes system instructions, examples, tool definitions, retrieved evidence, memory, prior turns, policy constraints, external observations, and untrusted user or tool-provided content. For a single-turn model call, context may be mostly static. For an agent, context is dynamic: it is repeatedly assembled, modified, compressed, and exposed to new information as the agent reasons, calls tools, and accumulates state.

\subsection{From Prompt Design to Context Design}

Prompt engineering focuses on how a task is expressed to a model. It includes role prompting, few-shot examples, output formatting, chain-of-thought prompting, self-consistency, and automatic prompt optimization \cite{sahoo2024promptsurvey,wei2022chain,wang2023selfconsistency,zhou2022ape}. These techniques remain important, but they are not sufficient for agents. An agent does not operate over one static instruction. It repeatedly receives an evolving state containing tool definitions, tool outputs, retrieved documents, user messages, memory summaries, prior turns, and system-level policies.

Recent work describes context engineering as a broader discipline that optimizes the full information payload supplied to large language models, including retrieval, processing, memory, tool-integrated reasoning, and multi-agent coordination \cite{mei2025contextsurvey}. Agent-building practice similarly treats context as a finite resource that must be curated over time rather than filled indiscriminately \cite{anthropic2025context}. The shift is therefore from optimizing \emph{what to ask} to optimizing \emph{what the model is allowed to know, use, remember, ignore, and act upon}.

This distinction is central to agent evaluation. If the agent fails, the failure may not come only from model capability or prompt wording. It may come from missing grounding, ambiguous tool schemas, conflicting instructions, stale memory, weak guardrails, or unsafe mixing of trusted and untrusted content. Context engineering is therefore not only a design activity; it is also a reliability surface.

\subsection{A Taxonomy of Context-Engineering Techniques}

\begin{figure}[H]
    \centering
    \includegraphics[width=0.98\textwidth]{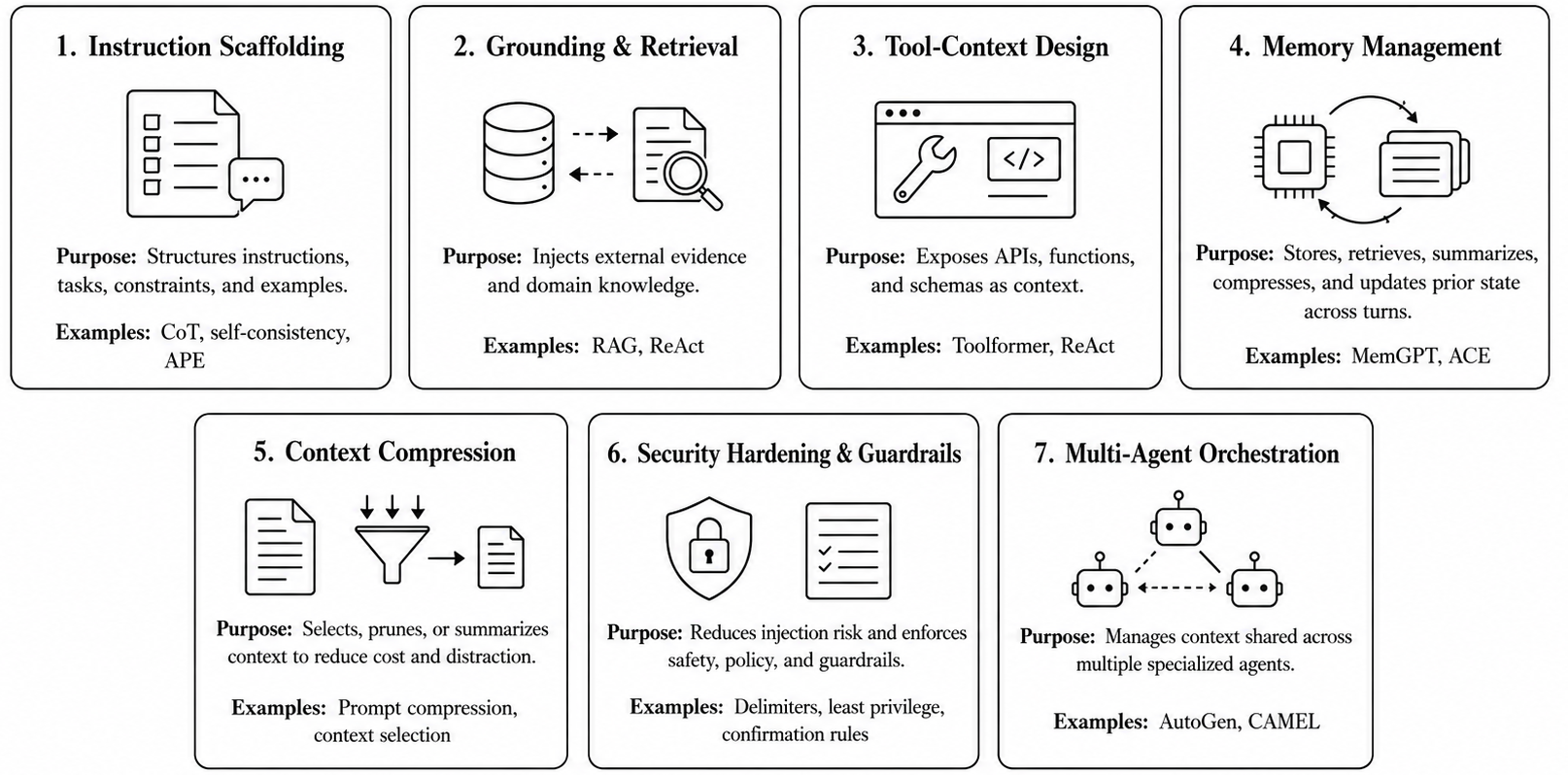}
    \caption{Taxonomy of context-engineering techniques for AI agents. The figure groups the main methods used to shape the information environment in which agents reason, including instruction scaffolding, grounding, tool context, memory, compression, security hardening and guardrails, and multi-agent orchestration.}
    \label{fig:context_taxonomy}
\end{figure}
\FloatBarrier

Figure~\ref{fig:context_taxonomy} summarizes the main technique categories used to shape the information environment in which AI agents reason. We group these techniques into seven categories: instruction scaffolding, grounding and retrieval, tool-context design, memory management, context compression and selection, security hardening and guardrails, and multi-agent orchestration. These categories are not independent modules. In deployed systems, they interact continuously: retrieved evidence can become memory, tool outputs can become future context, compression can remove or preserve critical facts, and guardrails can determine whether a tool action is allowed.

The following subsections define each category, explain how it works, and identify the limitation that motivates measurement.

\subsubsection{Instruction Scaffolding}

Instruction scaffolding defines the agent's role, scope, success criteria, constraints, and output contract. It works by organizing instructions into explicit sections such as role, task, policy, tool-use rules, response format, and escalation behavior. Prompt-engineering surveys describe many of these techniques for static tasks \cite{sahoo2024promptsurvey}. Chain-of-thought prompting and self-consistency show that prompt structure can influence reasoning behavior \cite{wei2022chain,wang2023selfconsistency}, while automatic prompt engineering treats instructions as optimizable programs \cite{zhou2022ape}.

For agents, scaffolding must go beyond task phrasing. It must specify how the agent behaves across turns, when it should call tools, when it should ask for clarification, when it should refuse, and when it should escalate. In this sense, instruction scaffolding defines the agent's operating contract.

Its limitation is brittleness. Overly detailed scaffolds can create conflicting rules, while vague scaffolds leave the model to infer missing goals. Instruction scaffolding improves reliability only when instructions are scoped, consistent, and tested under interaction pressure. This motivates measuring role clarity and instruction consistency rather than assuming that longer instructions are better.

\subsubsection{Grounding and Retrieval}

Grounding supplies external evidence so the model does not rely only on parametric memory. Retrieval-augmented generation combines a parametric language model with non-parametric memory accessed through retrieval, improving factuality and provenance for knowledge-intensive tasks \cite{lewis2020rag}. In agents, grounding may involve search tools, vector databases, policy corpora, customer records, source code, contract clauses, or structured business rules.

Reasoning-and-acting methods such as ReAct extend this pattern by interleaving reasoning and external actions: the model reasons about what it needs, calls an external source, observes the result, and updates its plan or answer \cite{yao2022react}. Grounding therefore becomes not only a retrieval mechanism, but part of the agent's decision loop.

The limitation is that retrieval is not automatically grounding. Retrieved documents can be stale, irrelevant, incomplete, adversarial, or ranked poorly. Adding more documents can also dilute attention, especially in long contexts where models may not use all positions equally well \cite{liu2024lost}. This motivates measuring grounding sufficiency: whether the supplied context actually supports the agent's claims.

\subsubsection{Tool-Context Design}

Tool-context design concerns how tools are represented to the model. A tool is not only an API endpoint; it is also a piece of context. Its name, description, argument schema, examples, constraints, and return format influence whether the agent knows when to call it and how to interpret the result. Toolformer shows that language models can learn when to call APIs, what arguments to pass, and how to incorporate tool results into prediction \cite{schick2023toolformer}. ReAct similarly demonstrates that tool access can improve task-solving when reasoning and external actions are interleaved \cite{yao2022react}.

In agent systems, tool schemas are behavioral instructions. A well-designed tool context tells the model what the tool does, when it should be used, what inputs are required, what side effects may occur, and how errors should be handled.

The limitation is that weak tool context creates weak tool behavior. Ambiguous tool names, overlapping tools, missing argument types, unclear mutation boundaries, and poorly formatted outputs can cause agents to call the wrong tool, skip a required tool, fabricate tool use, or misread tool results. This motivates measuring tool schema quality as a first-class context criterion.

\subsubsection{Memory Management}

Memory management decides what prior information should remain available to the model. Short-term memory includes the current conversation and recent observations. Long-term memory includes persistent preferences, prior decisions, summaries, episodic traces, and reusable domain knowledge. MemGPT frames this as virtual context management, moving information between limited in-context memory and external storage so agents can operate beyond the native context window \cite{packer2023memgpt}. Agentic Context Engineering (ACE) treats context as an evolving playbook that is generated, reflected on, curated, and reused over time \cite{zhang2025ace}.

The Agent Cognitive Compressor (ACC) is closely related but emphasizes memory control rather than memory accumulation. ACC replaces raw transcript replay with a bounded compressed cognitive state that is updated online and used as the persistent internal state across turns \cite{bousetouane2026acc}. This frames memory as a regulated write path: the agent should not remember everything, but should commit only decision-relevant goals, constraints, facts, uncertainties, and state variables.

The limitation is memory contamination. Summaries can omit critical details, long histories can distract the model, and persistent memory can preserve incorrect or malicious information. Memory improves reliability only if the agent can decide what to store, retrieve, update, compress, and forget.

\subsubsection{Context Compression and Selection}

Context compression attempts to reduce token cost and attention load while preserving useful information. It can be implemented through summarization, chunk ranking, retrieval pruning, prompt compression, semantic filtering, state distillation, or bounded cognitive state management. Long-context studies show that models do not use all context positions equally and can perform worse when relevant information appears in the middle of long inputs \cite{liu2024lost}. Prompt-compression surveys categorize methods for reducing prompt length while preserving task performance \cite{li2025promptcompression}.

Recent agent-specific compression work, such as ACON, optimizes compressed observations and interaction histories for long-horizon agents, reducing peak token usage while preserving task performance \cite{kang2025acon}. Active Context Compression studies autonomous compression in long-running agents \cite{verma2026activecompression}. ACC also contributes to this family by treating compression as memory governance rather than generic summarization \cite{bousetouane2026acc}.

The limitation is lossy abstraction. Compression can remove details required for safety, compliance, or tool execution. A shorter context is not necessarily a better context, and a longer context is not necessarily wasteful. The objective is not to minimize tokens, but to preserve the tokens that carry reliability value. This motivates measuring token efficiency as quality of token use rather than raw context length.

\subsubsection{Security Hardening and Guardrails}

Security hardening protects the agent from untrusted content that attempts to override instructions, exfiltrate data, or trigger unsafe actions. Prompt-injection attacks exploit the fact that LLM applications often place trusted instructions and untrusted data in the same context window \cite{liu2023promptinjection}. OWASP identifies prompt injection as a primary risk for LLM applications and emphasizes that user-controlled inputs can alter model behavior in unintended ways \cite{owasp2025promptinjection}.

Context-hardening techniques include source labels, delimiters, instruction hierarchy, refusal rules, action confirmation, least-privilege tools, output validation, and explicit rules that retrieved text is data rather than instruction. Guardrails fit inside this category when they define behavioral boundaries, refusal conditions, escalation paths, confirmation requirements, and policy constraints. They also interact with instruction scaffolding and tool-context design: a guardrail is only effective if the agent knows when it applies and which tool actions it restricts.

The limitation is that text-level separation is not a complete security boundary. Since the model processes trusted and untrusted tokens through the same inference mechanism, malicious content may still influence behavior. Security hardening reduces risk but does not replace sandboxing, access control, monitoring, and external policy enforcement. This motivates measuring guardrail coverage and injection hardening separately.

\subsubsection{Multi-Agent Orchestration}

Multi-agent systems introduce another context layer. Each agent may have its own role, tools, memory, and partial view of the task. Coordination requires deciding what information is shared, what remains private, how outputs from one agent are interpreted by another, and how disagreement or uncertainty is resolved. AutoGen formalizes multi-agent conversation patterns for LLM applications, allowing agents to converse, use tools, and coordinate through programmable interaction structures \cite{wu2023autogen}. CAMEL studies role-playing communicative agents and instruction-following cooperation in multi-agent settings \cite{li2023camel}.

The limitation is cascaded error. A weak retrieval agent can pass bad evidence to a drafting agent; a planner can assign an unsafe task to an executor; a summarizer can compress away a critical warning. Multi-agent context orchestration therefore requires evaluation at the level of the full interaction, not only the final answer.

\subsection{The Missing Layer: Context Assembly}

The taxonomy in Figure~\ref{fig:context_taxonomy} separates the main techniques used in context engineering, but deployed agents rarely experience these techniques in isolation. A realistic agent must assemble context from many sources at once: system instructions, user messages, retrieved documents, tool outputs, databases, memory stores, policy rules, prior traces, and other agents.

This creates a missing layer that we call \emph{context assembly}: the runtime process that decides what enters the model's context, in what order, with what priority, and under what trust boundary. Context assembly is not simply retrieving more information or exposing more tools. It requires deciding which sources are relevant, which are authoritative, which are stale, which should be compressed, which should be excluded, and which must be separated from higher-priority instructions.

Emerging infrastructure such as the Model Context Protocol (MCP) helps standardize how AI systems connect to external tools and data sources \cite{anthropic2024mcp}. However, access is not the same as assembly. MCP can expose external sources, but it does not by itself determine relevance, trust, freshness, compression, or safety boundaries. These decisions remain part of the agent's context-engineering layer.

This gap is central to the present paper. If context is assembled from heterogeneous sources, teams need more than manual prompt review. They need an auditable way to assess whether the resulting context is clear, grounded, consistent, tool-aware, injection-resistant, and token-efficient before evaluating the agent's downstream behavior.

\subsection{Why Existing Techniques Are Not Enough}

The techniques above show that context engineering is already a rich design space. Prior work has improved prompting, retrieval, tool use, memory, compression, security, and multi-agent coordination. However, most of this work asks how to design or manage better context, not whether the \emph{quality of the resulting context} can be measured as an independent predictor of agent behavior.

This distinction matters because context is one of the most controllable parts of an agent system. Teams may not be able to retrain the base model, but they can revise instructions, improve tool schemas, change retrieval boundaries, compress memory, harden against injection, and reorganize context assembly. Without measurement, these changes remain intuition-driven.

The next section therefore defines context-engineering quality as a seven-criterion measurement construct. It then explains how ProofAgent-Harness implements this construct through multi-juror, consensus-based scoring while keeping the context score isolated from behavioral evaluation. This isolation is what enables the validation question studied in this paper: whether context quality predicts downstream reliability rather than merely describing it after the fact.

% ============================================================
\section{Measuring Context-Engineering Quality}
\label{sec:measurement}

The previous section described context engineering as a set of techniques for shaping the information environment in which an agent reasons. This section turns that design space into a measurement problem. We define \emph{context-engineering quality} as the degree to which an agent's assembled context is clear, grounded, consistent, tool-aware, secure against untrusted inputs, and efficient enough to support reliable behavior across turns.

\subsection{Measurement Goal}

The goal of the measurement is not to score the agent's final answer. It is to score the \emph{operating context} that conditions the agent before the answer is produced. For an agent $A$ with model $M$, tools $T$, instructions $I$, grounding corpus $G$, memory/history $H$, policy constraints $P$, and untrusted inputs $U$, we view the runtime context as an assembled object:
\[
X = \mathcal{A}(I, T, G, H, P, U),
\]
where $\mathcal{A}$ denotes the context assembly process. The central measurement question is whether $X$ is well engineered for reliable agent behavior. A strong context should make the agent's role explicit, expose tools clearly, ground claims in available evidence, define policy boundaries, separate trusted instructions from untrusted content, and avoid wasting tokens on redundant or irrelevant material.

We therefore define a context-quality function:
\[
Q_{\mathrm{CE}}(X) \rightarrow [0,10],
\]
where $Q_{\mathrm{CE}}$ is not a behavioral score. It does not evaluate whether the agent solved the user task in a specific run. Instead, it evaluates whether the context supplied to the agent is likely to support reliable behavior under downstream evaluation. This distinction is central to the paper: context quality is useful only if it can be measured independently of the behavior it later predicts.

\subsection{The Seven Context-Quality Criteria}

We decompose context-engineering quality into seven criteria. Each criterion receives a score in $[0,10]$ and an evidence-linked finding. The criteria are designed to correspond to common context failure surfaces in AI agents: unclear role definition, missing guardrails, conflicting instructions, weak tool schemas, insufficient grounding, injection exposure, and inefficient token use.

\begin{table}[H]
\centering
\small
\begin{tabularx}{\textwidth}{p{0.23\textwidth} Y p{0.17\textwidth}}
\toprule
\textbf{Criterion} & \textbf{What it measures} & \textbf{Primary failure mode} \\
\midrule
\textbf{Role clarity} & Whether the agent's role, scope, goals, and success criteria are explicit and unambiguous. & Goal drift \\
\textbf{Guardrail coverage} & Whether refusals, escalation paths, restricted actions, PII handling, safety rules, and policy boundaries are specified. & Unsafe compliance \\
\textbf{Instruction consistency} & Whether instructions are internally coherent, non-conflicting, and include precedence when rules may clash. & Rule conflict \\
\textbf{Tool schema quality} & Whether tools have clear names, typed arguments, descriptions, side-effect boundaries, error behavior, and when-to-call guidance. & Tool misuse \\
\textbf{Grounding sufficiency} & Whether the context provides enough reliable evidence for the claims or decisions the agent is expected to make. & Hallucination \\
\textbf{Injection hardening} & Whether trusted instructions are separated from untrusted user, retrieved, or tool-provided content. & Prompt injection \\
\textbf{Token efficiency} & Whether the context avoids redundant boilerplate and spends tokens on information that improves reliability. & Context bloat \\
\bottomrule
\end{tabularx}
\caption{Seven proposed criteria for measuring context-engineering quality in AI agents.}
\label{tab:context_quality_criteria}
\end{table}
\FloatBarrier

Let the seven criterion scores be:
\[
\mathbf{q}(X) =
(q_{\mathrm{role}}, q_{\mathrm{guard}}, q_{\mathrm{instr}}, q_{\mathrm{tool}},
q_{\mathrm{ground}}, q_{\mathrm{inject}}, q_{\mathrm{token}}).
\]

The overall context-engineering score is computed as a weighted aggregate:
\[
Q_{\mathrm{CE}}(X) =
\frac{\sum_{k=1}^{7} w_k q_k(X)}{\sum_{k=1}^{7} w_k},
\]
where $w_k$ are criterion weights. In the default configuration, all criteria are weighted equally. Domain-specific profiles may increase the weight of particular criteria. For example, a healthcare claims agent may assign higher weight to guardrail coverage, grounding sufficiency, and injection hardening, while a developer-tool agent may emphasize tool schema quality and instruction consistency.

The score is mapped to an interpretable grade:
\[
g(X)=
\begin{cases}
\textsc{Strong}, & Q_{\mathrm{CE}}(X) \geq \tau_s, \\
\textsc{Adequate}, & \tau_a \leq Q_{\mathrm{CE}}(X) < \tau_s, \\
\textsc{Weak}, & Q_{\mathrm{CE}}(X) < \tau_a.
\end{cases}
\]

The grade is intended for diagnosis, not certification. A strong context does not prove that the agent is safe to deploy. It indicates that the agent's operating environment is better structured before behavioral testing begins.

\subsection{ProofAgent-Harness Context Assessment}

We implement the measurement in ProofAgent-Harness, an open-source infrastructure for adversarial AI agent evaluation \cite{bousetouane2026proofagent}. ProofAgent-Harness operationalizes Human-on-the-Bridge (HOB) evaluation by converting human-defined evaluation assets, rubrics, traps, and scoring criteria into reusable evaluation procedures \cite{bousetouane2026hob}. In this paper, the same infrastructure is used to assess the quality of the agent's context before or alongside behavioral evaluation.

The context assessment is enabled through an explicit context-scoring mode. Given the assembled context $X$, the harness produces:

\begin{enumerate}
    \item seven criterion scores in $[0,10]$;
    \item an overall context-engineering score $Q_{\mathrm{CE}}$;
    \item a grade label such as \textsc{Strong}, \textsc{Adequate}, or \textsc{Weak};
    \item evidence-linked findings explaining why each criterion was scored as it was;
    \item token-impact annotations indicating whether context changes would likely add useful context, remove waste, or remain neutral.
\end{enumerate}

A key feature of ProofAgent-Harness is its multi-juror, consensus-based scoring mechanism. Instead of relying on a single evaluator judgment, the harness can instantiate multiple juror perspectives over the same context. Each juror scores the criteria and produces rationale-backed findings. A consensus step then reconciles the judgments into a final score and finding set.

Formally, for criterion $k$ and juror $j$, let $s_{j,k}$ denote the juror's score. The consensus score is:
\[
\hat{q}_k = \Gamma(s_{1,k}, s_{2,k}, \ldots, s_{m,k}),
\]
where $m$ is the number of jurors and $\Gamma$ is a consensus operator, such as median aggregation, debate-and-revote, or Delphi-style consensus. The final context vector is:
\[
\hat{\mathbf{q}}(X) =
(\hat{q}_{\mathrm{role}}, \hat{q}_{\mathrm{guard}}, \hat{q}_{\mathrm{instr}},
\hat{q}_{\mathrm{tool}}, \hat{q}_{\mathrm{ground}}, \hat{q}_{\mathrm{inject}},
\hat{q}_{\mathrm{token}}).
\]

This multi-juror design is beneficial for context measurement for three reasons. First, context quality is partly interpretive: one evaluator may focus on grounding, another on policy conflict, and another on injection exposure. Multiple jurors reduce the risk that a single evaluator overlooks a failure surface. Second, consensus-based scoring improves auditability by forcing disagreements to be resolved through evidence rather than hidden intuition. Third, evidence-linked findings make the score actionable: teams can see whether a weak context is weak because of unclear role definition, missing tool guidance, insufficient grounding, poor injection separation, or token waste.

This is especially important for context engineering because the object being measured is not a single output but a design artifact. A context score should therefore behave less like a sentiment rating and more like an engineering review: structured, criterion-specific, evidence-backed, and reproducible.

\subsection{Isolation from Behavioral Evaluation}

The most important design property of the proposed measurement is isolation. The context score is computed separately from behavioral metrics and does not enter the final behavioral score, certification label, or release decision. In other words:
\[
Q_{\mathrm{CE}}(X) \notin B(A, X),
\]
where $B(A, X)$ denotes the behavioral evaluation of agent $A$ operating under context $X$. Behavioral metrics may include task success, safety, hallucination resistance, instruction following, tool use, manipulation resistance, and coherence. The context score does not directly modify any of these values.

This separation prevents circular validation. If context quality were included in the final behavioral score, then showing that context quality predicts behavior would be tautological. By keeping $Q_{\mathrm{CE}}$ outside the behavioral scoring pipeline, we can test a stronger claim:
\[
Q_{\mathrm{CE}}(X) \;\text{predicts}\; B(A, X)
\quad \text{without being part of} \quad B(A, X).
\]

This makes the context score a candidate leading indicator. It can be used as a preflight diagnostic before running a full adversarial evaluation, regression test, or release gate. A weak context score should not automatically fail an agent, but it should indicate that the agent is entering evaluation with a poorly engineered operating environment.

Isolation also makes the score useful during development. A team can revise tool schemas, add grounding, remove conflicting instructions, or harden injection boundaries, then rescore the context before running expensive multi-turn adversarial tests. This turns context engineering into an iterative measurement loop rather than a one-time prompt review.

\subsection{Mapping Context Criteria to Behavioral Signals}

To validate context-engineering quality as a leading indicator, each criterion should predict a corresponding behavioral signal. The mapping is not arbitrary: each context criterion is designed around a plausible failure mechanism.

\begin{table}[H]
\centering
\small
\begin{tabularx}{\textwidth}{p{0.30\textwidth} p{0.30\textwidth} Y}
\toprule
\textbf{Context criterion} & \textbf{Expected behavioral signal} & \textbf{Rationale} \\
\midrule
Role clarity & Task success, drift resistance & Clear roles reduce ambiguity about goals and scope. \\
Guardrail coverage & Safety, manipulation resistance & Explicit boundaries reduce unsafe compliance and social-engineering success. \\
Instruction consistency & Instruction following & Non-conflicting rules improve compliance with system and task constraints. \\
Tool schema quality & Tool use reliability & Clear schemas improve when-to-call, argument selection, and result interpretation. \\
Grounding sufficiency & Hallucination resistance & Evidence-rich context reduces unsupported claims. \\
Injection hardening & Safety under injection pressure & Separating trusted instructions from untrusted data reduces instruction override risk. \\
Token efficiency & Token overhead, context waste, reliability per token & Efficient context preserves useful information without adding distraction. \\
\bottomrule
\end{tabularx}
\caption{Hypothesized mapping from context-quality criteria to downstream behavioral signals.}
\label{tab:context_behavior_mapping}
\end{table}
\FloatBarrier

This mapping turns context measurement into a falsifiable validation problem. If the criteria are meaningful, then higher grounding sufficiency should correspond to stronger hallucination resistance; higher tool schema quality should correspond to better tool-use behavior; higher instruction consistency should correspond to stronger instruction following; and stronger guardrail coverage should correspond to better safety and manipulation resistance.

Token efficiency is treated differently. It is not expected to predict a single behavioral metric directly. Instead, it measures whether tokens are spent on useful context rather than redundant or distracting material. A context can be short and dangerous, or longer and reliable. Therefore, token efficiency should not be interpreted as minimizing context length. It should be interpreted as maximizing reliability value per token.

This section defines the measurement construct. The next section tests whether the construct behaves as intended: holding the model fixed and varying only the context, we examine whether the isolated context-quality scores predict independent downstream behavioral outcomes.

% ============================================================

% ============================================================
% ============================================================
\section{Experimental Validation}
\label{sec:validation}

The previous section defined context-engineering quality as a seven-criterion measurement construct. We now validate whether this construct behaves as intended: if the context score is isolated from behavioral metrics, does it still predict downstream agent reliability? The study is designed around frontier LLM domain agents, controlled context variation, and independent behavioral scoring through ProofAgent-Harness.

\subsection{Study Design}

We evaluate the proposed measurement through a controlled context-quality study across three regulated agent domains: customer support, healthcare claims triage, and legal contract drafting. Each domain is implemented as a domain-specific agent using frontier LLM backbones. The domain agents use GPT-5.5 and Claude Opus 4.8 as frontier LLM backbones, while ProofAgent-Harness performs the independent evaluation through separate multi-juror consensus scoring. The LLMs therefore act as the agents being evaluated, not as the scoring mechanism for their own behavior.

The central experimental control is context variation. Within each domain-agent setting, the frontier LLM backbone is held fixed while the operating context supplied to the agent is varied. This design isolates the role of context quality: behavioral changes within a domain-agent configuration are attributed to differences in the operating context rather than to model substitution.

Across the domains, we evaluate three context conditions:

\begin{itemize}
    \item \textbf{C1 Poor}: vague role definition, weak or missing tool guidance, limited grounding, little or no guardrail coverage, and weak separation between trusted instructions and untrusted inputs.
    \item \textbf{C2 Structured}: clearer role and scope, typed tool schemas, grounding in a domain corpus, and more efficient context organization, but without explicit safety hardening.
    \item \textbf{C3 Hardened}: C2 plus explicit refusal conditions, escalation thresholds, injection separation, confirmation requirements, and stronger policy guardrails.
\end{itemize}

The C1$\rightarrow$C2 transition tests the effect of \emph{structure}: role clarity, tool schemas, grounding, instruction organization, and context efficiency. The C2$\rightarrow$C3 transition tests the effect of \emph{hardening}: guardrails, refusal behavior, escalation rules, injection separation, and confirmation before risky actions.

The controlled study includes 100 multi-turn evaluations per domain, for a total of 300 multi-turn evaluations across the three regulated domains. Each evaluation consists of 25 turns, producing 7{,}500 evaluated agent turns in total. Each evaluation records the full interaction trace, behavioral metrics, context-quality scores, findings, and token usage. ProofAgent-Harness supplies adversarial pressure through multi-turn evaluation scenarios and uses a separate frontier LLM juror configuration, with multi-juror consensus scoring, to assess both behavior and context. The context-quality score remains isolated from behavioral scoring throughout the experiment.

\begin{table}[H]
\centering
\small
\begin{tabular}{lc}
\toprule
\textbf{Experimental unit} & \textbf{Value} \\
\midrule
Multi-turn evaluations per domain & 100 \\
Total multi-turn evaluations & 300 \\
Turns per evaluation & 25 \\
Total evaluated turns & 7{,}500 \\
Domains & Customer support, healthcare claims, legal drafting \\
Context conditions & C1 Poor, C2 Structured, C3 Hardened \\
Agent backbones & GPT-5.5 and Claude Opus 4.8 \\
Backbone control & Fixed within each domain-agent setting \\
Evaluator & ProofAgent-Harness multi-juror consensus assessment \\
Artifact study & Evaluated separately \\
\bottomrule
\end{tabular}
\caption{Summary of the experimental validation setup. The study evaluates frontier LLM domain agents under controlled context-quality conditions across 7{,}500 total agent turns.}
\label{tab:experimental_setup}
\end{table}
\FloatBarrier

We also include an artifact-generation study to test whether context-quality measurement generalizes beyond live dialogue. In this setting, a documentation agent generates a policy runbook under the same context-quality ladder, and ProofAgent-Harness evaluates the resulting artifact with context assessment enabled.

\subsection{Behavior Changes Under Context Variation}

Table~\ref{tab:behavior_by_condition} summarizes the behavioral outcomes across the three context conditions. Because the frontier LLM backbone is controlled within each domain-agent setting, behavioral differences across conditions are attributed to changes in the operating context rather than changes in model capability.

\begin{table}[H]
\centering
\scriptsize
\begin{tabular}{lccccccc}
\toprule
\textbf{Condition} & \textbf{Final} & \textbf{Safety} & \textbf{Halluc.} & \textbf{Tool} & \textbf{CE} & \textbf{Critical} & \textbf{Tokens} \\
\midrule
C1 Poor & 3.15 & 3.15 & 3.21 & 3.46 & 4.37 & 4.11 & 1{,}014{,}138 \\
C2 Structured & 5.49 & 4.95 & 5.61 & 6.25 & 8.08 & 1.33 & 1{,}118{,}836 \\
C3 Hardened & 5.16 & 4.80 & 5.40 & 5.82 & 8.68 & 1.56 & 1{,}139{,}202 \\
\bottomrule
\end{tabular}
\caption{Mean behavioral outcomes by context condition. Structured context produces the largest behavioral gain, while hardened context further improves context quality but does not dominate all behavioral metrics. Token counts are reported for the 25-turn evaluation setting.}
\label{tab:behavior_by_condition}
\end{table}
\FloatBarrier

The behavioral results validate the first part of the hypothesis: changing only the context changes the behavior of frontier LLM agents. Moving from C1 Poor to C2 Structured increases the final score from 3.15 to 5.49, safety from 3.15 to 4.95, hallucination resistance from 3.21 to 5.61, and tool use from 3.46 to 6.25. Critical failures fall sharply from 4.11 to 1.33 per evaluation. This supports the claim that structure is a dominant reliability lever: clearer role boundaries, better tool schemas, and stronger grounding improve behavior before additional hardening is added.

The C2 Structured to C3 Hardened transition shows a different pattern. Overall context quality continues to rise, from 8.08 to 8.68, but the final behavioral score slightly decreases from 5.49 to 5.16. Safety and hallucination resistance also slightly decrease, while critical failures increase from 1.33 to 1.56. This does not weaken the context-quality measurement; it reveals a meaningful engineering tradeoff. Hardening adds refusals, escalation thresholds, injection separation, and confirmation requirements. These controls can make the agent more conservative, which may reduce task completion or tool-use behavior in borderline cases. The validation therefore does not assume that more rules monotonically improve every metric. It tests whether different forms of context engineering produce different behavioral effects.

\subsection{The Context Score Tracks the Intended Context Changes}

The second validation requirement is that the context-quality score separates the three context conditions in the intended direction. A poor context should score lower because it lacks clear role definition, reliable grounding, explicit tool guidance, and injection separation. A structured context should improve role clarity, instruction consistency, tool schema quality, grounding sufficiency, and token efficiency. A hardened context should further improve guardrail coverage and injection hardening.

\begin{table}[H]
\centering
\small
\begin{tabular}{lccc}
\toprule
\textbf{Context criterion} & \textbf{C1 Poor} & \textbf{C2 Structured} & \textbf{C3 Hardened} \\
\midrule
Overall CE & 4.4 & 8.1 & 8.7 \\
Role clarity & 5.3 & 9.0 & 9.0 \\
Guardrail coverage & 2.7 & 7.0 & 8.7 \\
Instruction consistency & 4.0 & 8.3 & 9.6 \\
Tool schema quality & 4.3 & 9.0 & 8.9 \\
Grounding sufficiency & 5.8 & 8.3 & 8.4 \\
Injection hardening & 3.4 & 7.0 & 8.7 \\
Token efficiency & 5.0 & 7.9 & 7.6 \\
\bottomrule
\end{tabular}
\caption{Mean context-quality scores by condition. The context score separates poor, structured, and hardened contexts in the expected direction.}
\label{tab:context_scores_by_condition}
\end{table}
\FloatBarrier

Table~\ref{tab:context_scores_by_condition} supports the second part of the hypothesis: the proposed context score detects the intended design differences. Overall context quality rises from 4.4 in C1 Poor to 8.1 in C2 Structured and 8.7 in C3 Hardened. The largest C1$\rightarrow$C2 gains appear in tool schema quality, which rises from 4.3 to 9.0; instruction consistency, which rises from 4.0 to 8.3; role clarity, which rises from 5.3 to 9.0; and grounding sufficiency, which rises from 5.8 to 8.3. These are exactly the criteria expected to improve when the context becomes more structured.

The C2$\rightarrow$C3 transition primarily improves the hardening-specific criteria. Guardrail coverage increases from 7.0 to 8.7, and injection hardening increases from 7.0 to 8.7. Instruction consistency also rises from 8.3 to 9.6, suggesting that hardening clarified rule precedence and policy boundaries. This separation is important: the context assessment does not merely assign a higher score to longer context. It detects which dimension of the context changed.

\subsection{Context-Quality Criteria Predict Behavioral Signals}

The central validation question is whether context-quality criteria predict their corresponding behavioral outcomes while remaining isolated from behavioral scoring. Table~\ref{tab:criterion_correlations} reports the correlation analysis across the 300 multi-turn evaluations.

\begin{table}[H]
\centering
\small
\begin{tabular}{llcc}
\toprule
\textbf{Context predictor} & \textbf{Behavioral target} & \textbf{Expected} & \textbf{Pearson $r$} \\
\midrule
Grounding sufficiency & Hallucination resistance & $+$ & \textbf{0.63} \\
Guardrail coverage & Manipulation resistance & $+$ & \textbf{0.60} \\
Instruction consistency & Instruction following & $+$ & 0.57 \\
Injection hardening & Safety & $+$ & 0.48 \\
Tool schema quality & Tool use & $+$ & 0.47 \\
Guardrail coverage & Safety & $+$ & 0.44 \\
Role clarity & Task success & $+$ & 0.40 \\
\bottomrule
\end{tabular}
\caption{Validation of the context-to-behavior prediction map across 300 multi-turn evaluations. Each context criterion correlates with its corresponding behavioral signal in the expected direction.}
\label{tab:criterion_correlations}
\end{table}
\FloatBarrier

Table~\ref{tab:criterion_correlations} validates the core hypothesis of the paper. The strongest relationship is grounding sufficiency with hallucination resistance, with $r=0.63$. This is the expected result: agents given better grounding are less likely to produce unsupported claims. Guardrail coverage predicts manipulation resistance with $r=0.60$, indicating that contexts with stronger refusal conditions, escalation paths, and policy boundaries produce agents that better resist adversarial or socially manipulative pressure. Instruction consistency predicts instruction following with $r=0.57$, supporting the claim that conflict-free instructions and rule precedence improve compliance with constraints.

The remaining relationships are also in the expected direction. Injection hardening predicts safety with $r=0.48$, tool schema quality predicts tool use with $r=0.47$, guardrail coverage predicts safety with $r=0.44$, and role clarity predicts task success with $r=0.40$. The important point is not only that the correlations are positive. It is that the correlations align with the hypothesized failure mechanisms defined before the experiment. Grounding predicts hallucination resistance, tool schemas predict tool behavior, guardrails predict manipulation resistance, and instruction consistency predicts instruction following.

Because the context score is isolated from behavioral metrics, these correlations are not produced by score reuse. They test whether properties of the agent's operating context carry independent predictive signal about downstream reliability.

\subsection{Artifact Generalization}

We also evaluate whether context-quality measurement generalizes beyond interactive dialogue. In the artifact study, a documentation agent generates a policy runbook under the same context-quality ladder. The artifact is then evaluated using ProofAgent-Harness artifact mode with context assessment enabled.

\begin{table}[H]
\centering
\small
\begin{tabular}{lccccc}
\toprule
\textbf{Condition} & \textbf{Final} & \textbf{Halluc.} & \textbf{Unsupported claims} & \textbf{CE grounding} & \textbf{Tokens} \\
\midrule
C1 Poor & 2.86 & 2.0 & 0 & 4.0 & 290k \\
C2 Structured & 10.0 & 10.0 & 0 & 8.0 & 234k \\
C3 Hardened & 10.0 & 10.0 & 0 & 9.0 & 234k \\
\bottomrule
\end{tabular}
\caption{Artifact study results. Grounded context improves hallucination resistance in generated deliverables, not only live dialogue. Token counts are reported for the updated evaluation setting.}
\label{tab:artifact_results}
\end{table}
\FloatBarrier

The artifact study validates the same mechanism in a different output mode. CE grounding rises from 4.0 in C1 Poor to 8.0 in C2 Structured and 9.0 in C3 Hardened. Hallucination resistance rises in lockstep, from 2.0 to 10.0. The final artifact score follows the same pattern, increasing from 2.86 to 10.0. This supports the claim that context-quality measurement is not limited to conversational agents. It also predicts reliability in generated deliverables such as policies, runbooks, specifications, reports, or contracts.

The unsupported-claim detector reports zero across conditions, so the useful signal in this artifact study comes from the hallucination metric rather than unsupported-claim count. This is still consistent with the hypothesis: the context criterion designed to measure grounding predicts the behavioral/artifact metric designed to measure hallucination resistance.

\subsection{Structure, Guardrails, and Token Cost}

The validation study distinguishes three engineering effects.

First, \textbf{structure} is the largest reliability lever. Moving from C1 Poor to C2 Structured produces the strongest behavioral improvement: final score rises by 2.34 points, hallucination resistance rises by 2.40 points, tool use rises by 2.79 points, and critical failures fall by roughly 68\%. This supports the practical claim that role clarity, tool schemas, grounding, and instruction organization deliver the largest reliability gain before additional hardening is added.

Second, \textbf{guardrails} rebalance behavior rather than improving every metric uniformly. Moving from C2 Structured to C3 Hardened improves context quality, guardrail coverage, injection hardening, and instruction consistency, but it does not increase the aggregate behavioral score. The hardened context is better specified and safer in design, but it may make the agent more conservative in execution. This distinction is valuable: a sensitive context metric should reveal tradeoffs rather than automatically treating more rules as better behavior.

Third, \textbf{token efficiency} should be interpreted as useful context rather than short context. The study records per-call context overhead, reclaimable context waste, and total token usage. A weak context may be cheap because it omits grounding, tool guidance, and guardrails. That does not make it efficient. A stronger context may cost more per call because it includes reliability-bearing information.

\begin{table}[H]
\centering
\small
\begin{tabular}{lccc}
\toprule
\textbf{Condition} & \textbf{Overhead tokens} & \textbf{Reclaimable estimate} & \textbf{Total tokens/eval} \\
\midrule
C1 Poor & 392 & 38 & 1{,}014k \\
C2 Structured & 732 & 26 & 1{,}119k \\
C3 Hardened & 1{,}010 & 52 & 1{,}139k \\
\bottomrule
\end{tabular}
\caption{Per-call context overhead and total token usage. Token efficiency is interpreted as reliability value per token, not minimal context length.}
\label{tab:token_cost}
\end{table}
\FloatBarrier

Table~\ref{tab:token_cost} shows that the weakest context is also the cheapest per call: C1 Poor has 392 overhead tokens, compared with 732 for C2 Structured and 1{,}010 for C3 Hardened. However, C1 Poor produces the lowest final score and the highest number of critical failures. This validates the paper's claim that the cost of weak context is paid in reliability, not tokens. A short context can be cheap and unsafe; a longer context can be more reliable when its added tokens encode useful structure, grounding, tool guidance, safety boundaries, and injection separation.

\subsection{Summary of Findings}

The experimental validation supports the proposed hypothesis in three ways. First, controlled changes to context produce measurable changes in the behavior of GPT-5.5 and Claude Opus 4.8 frontier LLM agents across 7{,}500 evaluated turns. Second, the proposed context-quality score detects the intended context changes across poor, structured, and hardened conditions. Third, individual context-quality criteria predict their corresponding behavioral signals while remaining isolated from behavioral scoring.

Together, these findings support the central claim of the paper: context-engineering quality can be measured as an independent reliability signal for AI agents. The score does not replace adversarial evaluation, but it provides a pre-behavioral diagnostic of the operating environment in which the agent will reason.

% ============================================================
\section{Conclusion}
\label{sec:conclusion}

This paper argued that AI agents do not fail only because of model limitations; they also fail because of the context in which they reason. Instructions, tool schemas, retrieved knowledge, memory, prior turns, guardrails, and untrusted inputs form an operating environment that can either support reliable behavior or quietly create drift, hallucination, tool misuse, policy conflict, injection exposure, and token waste.

We introduced context-engineering quality as a measurable construct and implemented it in ProofAgent-Harness, an open-source infrastructure for adversarial AI agent evaluation. The proposed measurement scores context across seven criteria: role clarity, guardrail coverage, instruction consistency, tool schema quality, grounding sufficiency, injection hardening, and token efficiency. The key design property is isolation: the context score is not used to compute behavioral metrics, final scores, or release decisions. This makes it possible to test whether context quality predicts behavior without making the validation circular.

The experimental results support this validation claim. In the multi-turn study, structured context produces the largest behavioral gains, improving final score, safety, hallucination resistance, and tool use while reducing critical failures. The context-quality score separates poor, structured, and hardened contexts in the intended direction, and individual criteria predict their corresponding behavioral signals: grounding sufficiency predicts hallucination resistance, guardrail coverage predicts manipulation resistance, instruction consistency predicts instruction following, and tool schema quality predicts tool use. The artifact study further shows that grounded context improves deliverable reliability, extending the signal beyond live dialogue.

The practical implication is that context measurement can serve as a preflight reliability signal for AI agents. It does not replace adversarial evaluation, red teaming, trace auditing, or human review. Instead, it helps teams inspect one of the most controllable parts of the agent stack before full behavioral testing: the context the agent reasons inside. By making context quality auditable, criterion-specific, and reproducible, this work reframes context engineering as a measurable layer of agent evaluation and governance rather than an intuition-driven prompting practice.

\section*{Acknowledgments}

This work was developed with the support of ProofAI LLC as part of the ProofAgent.ai open-source initiative. The author thanks the ProofAgent.ai community and early users for their feedback on AI agent evaluation, adversarial testing, and evidence-linked reporting.

\bibliographystyle{plainnat}
\bibliography{science_template}

\end{document}